\documentclass[letterpaper, 10 pt, conference]{ieeeconf}  

\IEEEoverridecommandlockouts                              
\usepackage{floatrow}
\newfloatcommand{capbtabbox}{table}[][\FBwidth]
\usepackage{graphicx}
\usepackage{comment}
\usepackage{amsmath,amssymb} 
\usepackage{color}
\usepackage{hyperref}
\usepackage{subfigure} 
\usepackage{booktabs} 

\usepackage{graphics} 
\usepackage{amsmath} 
\usepackage{amssymb}  
\usepackage{hyperref}
\usepackage{cite}
\usepackage{subfigure}
\usepackage{graphicx}
\usepackage{setspace}
\usepackage{multirow,makecell}
\usepackage{xcolor}
\usepackage{wasysym}
\title{\LARGE \bf Semantic Dense Reconstruction with Consistent Scene Segments}  

\author{Yingcai Wan$^{1,4*}$, Yanyan Li$^{2,4*}$, Yingxuan You$^{3}$, Cheng Guo$^{1,4}$, Lijin Fang$^{1}$ and Federico Tombari$^{2,5}$ 
\thanks{$^{1}$Dept. Faculty of Robot Science and engineering, Northeastern University, Shenyan, China.
        {\tt\small (wan.yc.meta,guochengrobot)@gmail.com, ljfang@mail.neu.edu.cn}}%
\thanks{$^{2}$Dept. Computer Science, Technical University of Munich, Munich, Germany.
        {\tt\small (yanyan.li,federico.tombari)@tum.de}}%
\thanks{$^{3}$Key Laboratory of Machine Perception, Peking university, Beijing, China.
        {\tt\small youyx@stu.pku.edu.cn}}%
\thanks{$^4$ MetaSpatie Tech. $^5$ Google Inc. *authors are with equal contributions.}          
}

\begin{document}

\maketitle
\thispagestyle{empty}
\pagestyle{empty}

\begin{abstract}
In this paper, a method for dense semantic 3D scene reconstruction from an RGB-D sequence is proposed to solve high-level scene understanding tasks. 
First, each RGB-D pair is consistently segmented into 2D semantic maps based on a camera tracking backbone that propagates objects' labels with high probabilities from full scans to corresponding ones of partial views. 
Then a dense 3D mesh model of an unknown environment is incrementally generated from the input RGB-D sequence.  
Benefiting from 2D consistent semantic segments and the 3D model, a novel semantic projection block (SP-Block) is proposed to extract deep feature volumes from 2D segments of different views. Moreover, the semantic volumes are fused into deep volumes from a point cloud encoder to make the final semantic segmentation.   
Extensive experimental evaluations on public datasets show that our system achieves accurate 3D dense reconstruction and state-of-the-art semantic prediction performances simultaneously.
\end{abstract}

\section{Introduction}
Scene understanding systems support robots and smart devices to interact intelligently with unknown environments, which aim to provide spatial and semantic information around 3D instances of scenes. Commonly, the majority of 3D semantic segmentation methods~\cite{qi2017pointnet++,8594391,hu2021bidirectional} rely on the assumption of having available as input a complete and accurate 3D model, which is hard to obtain under realistic settings.  
Therefore, how to provide a complete system for this 3D scene understanding task from RGB-D images is still open in the community of computer vision.

\begin{figure}
    \centering
    \subfigure[RGB-D input]{
    \includegraphics[width=0.47\linewidth]{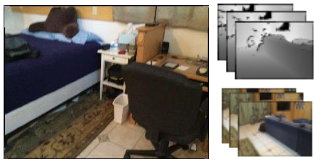}}
    \subfigure[Consistent masks]{
    \includegraphics[width=0.47\linewidth]{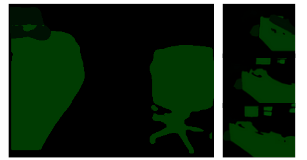}}
    \subfigure[Dense reconstruction]{
    \includegraphics[width=0.47\linewidth]{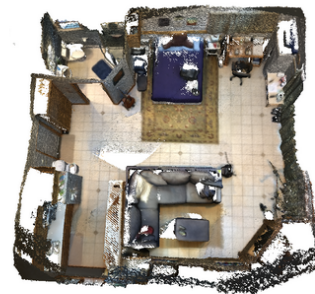}}
    \subfigure[Semantic instance map]{
    \includegraphics[width=0.47\linewidth]{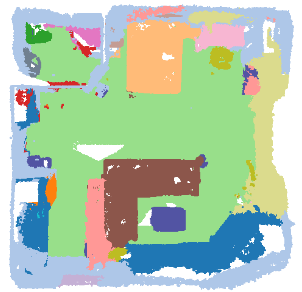}}
    \caption{Input (a) and Outputs (b, c, d) of the system. Consistent semantic masks (b) and a dense mesh model (c) support our network to predict semantic instances (d). }
    \label{fig:teaser}
\end{figure}


Given a single RGB image, 2D semantic segmentation algorithms~\cite{vuola2019mask,2017Faster,yolact-iccv2019,yolact-plus-tpami2020} can obtain impressive performance. These methods, however, are sensitive to viewpoint changes and also suffer from inconsistent predictions across different views. 
To solve these issues, deep neural networks have been applied to 3D semantic segmentation tasks based on point clouds, 
such as PointNet++\cite{qi2017pointnet++}, MCCNN~\cite{2018Monte} and MinkowskiNet~\cite{choy20194d,tang2020searching}. 
The primary focus of these works lies on the analysis of point clouds from complete dense reconstruction results, while the geometric and semantic details captured in the process of reconstruction are ignored. 
Recently, several 2D-3D joint end-to-end  methods~\cite{dai20183dmv,hou20193d} combining 3D knowledge (geometry and shape) with 2D detailed information (texture and color) were proposed to improve the accuracy of semantic segmentation tasks.
Following the 2D-3D fusion strategy, BPNet~\cite{hu2021bidirectional} proposes a bidirectional projection module to improve both 2D and 3D semantic segmentation performance from 2D RGB images and point clouds. 
However, these attempts are performed without fully leveraging the complementay 2D information. Furthermore, they assume that the input 3D scenes are already completely reconstructed at high quality, rather than building dense models from real RGB-D images. 

%

To build a complete scene semantic estimation pipeline from images, the pioneering benchmark, ScanNet~\cite{dai2017scannet} and Matterport3D~\cite{2017Matterport3D} achieve dense reconstruction and semantic segmentation at the same time. Different from them, our tracking and dense mapping part are implemented on a CPU.  
SceneGraphFusion~\cite{2021SceneGraphFusion} also explores to start from images and obtains 3D semantic models. Nevertheless, the system makes use of ground truth camera poses and build an instance point cloud by using a geometric instance segmentation method~\cite{tateno2015real}, which limits the prediction performance when those instances are not fused accurately. 

Compared to the existing 3D segmentation networks~\cite{choy20194d,hu2021bidirectional} and scene graph generation approach~\cite{2021SceneGraphFusion}, our dense semantic reconstruction method has a more complete function that adopts continuous RGB-D frames and outputs dense semantic 3D models. 
Based on 2D segmentation algorithms~\cite{yolact-iccv2019,yolact-plus-tpami2020} and our camera trackers~\cite{yunus2021manhattanslam,Li2021PlanarSLAM}, we maintain a semantic sparse map that saves the probability of each object. Since semantic predictions from partial views are not as reliable as full ones', the semantic sparse map is used to correct wrong 2D segments existed in bad cases.  
After obtaining camera poses, consistent 2D semantic masks and a dense 3D reconstruction model, the proposed SP-Block is responsible to extracts the multi-scale features from 2D consistent object segments and project channels of deep features into volumes that are fused with those extracted from the encoder of MinkowskiNet~\cite{choy20194d} after the domain transformation (DoT) operation as shown in Figure~\ref{fig:FSmodule}. Compared with MinkowskiNet~\cite{choy20194d} and BPNet~\cite{hu2021bidirectional}, our 2D semantic masks of different views provide accurate 2D objects' regions that make traditional 3D semantic predictions more accurate.    
\begin{figure*}
    \centering
    \includegraphics[width=\linewidth]{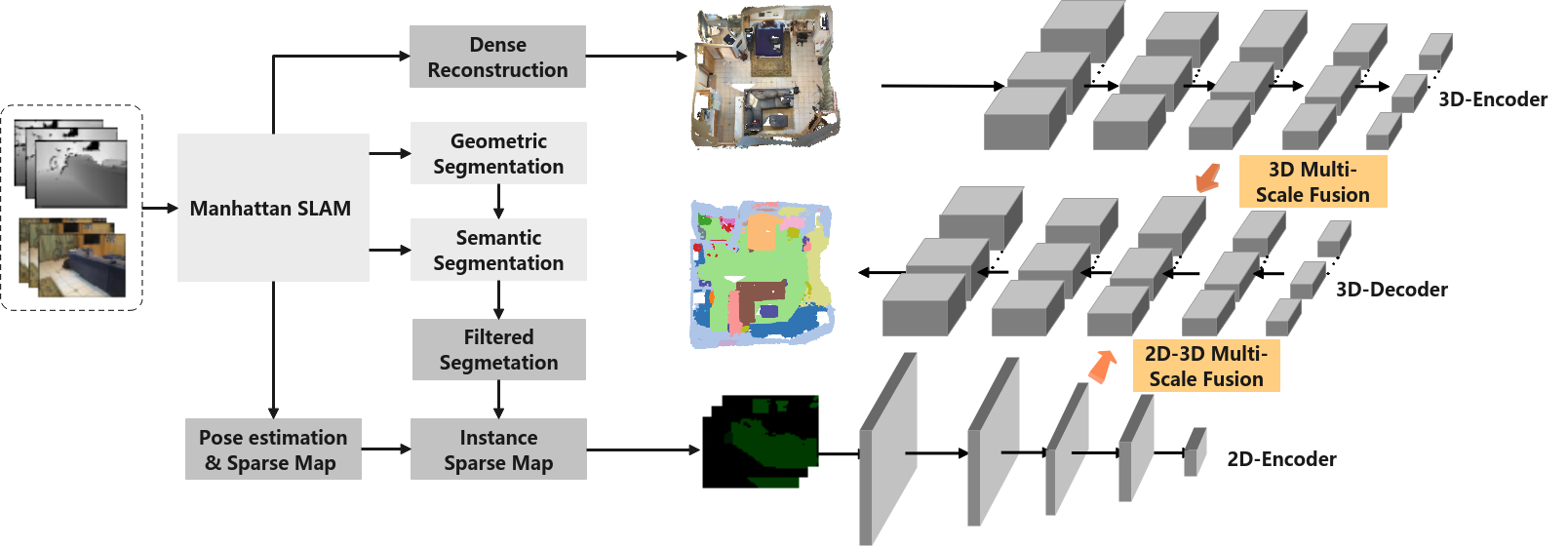}
    \caption{Pipeline of the proposed system that is fed by sequential RGB-D pairs and generate 2D consistent semantic masks, a dense mesh model and a 3D semantic segmentation result.} 
    \label{fig:pipeline}
\end{figure*}
%
%
The contributions of this paper are summarized as follows,
\begin{itemize}
    \item The 2D consistent object prediction strategy based on 2D segments and a sparse semantic map is proposed to achieve accurate and consistent 2D semantic masks.
    \item SP-Block is built to extract and project multi-scale deep features from 2D semantic map, which are transformed to the feature domain from point clouds by the DoT operation. 
    \item The proposed architecture containing tracking, meshing, and segmentation modules extends traditional SLAM methods to a multi-task scene understanding system.  
\end{itemize}

\section{Related work}
\subsection{Object instance segmentation}
Early 2D semantic segmentation methods, including Faster R-CNN~\cite{2017Faster}, Mask R-CNN~\cite{vuola2019mask}, make instance mask predictions before object semantic recognition. More recent networks~\cite{tian2021boxinst,carion2020end,zhu2019feature} are anchor-based approaches that predict boxes' offsets relative to a collection of fixed boxes. 
Although semantic instance segmentation has achieved reliable results, more and more segmentation tasks have put forward requirements for efficiency.
%
YOLACT~\cite{yolact-iccv2019} is the real-time (more than $30$ fps) instance segmentation algorithm that is updated as YOLACT++~\cite{yolact-plus-tpami2020} by incorporating deformable convolutions into the backbone network. 
However, those approaches focus on the single image processing topics, leading to inconsistent scene interpretation issues due to illumination changes, occlusions and other variations over time. 
To solve the problem, video-based instance segmentation (VIS)~\cite{wang2021end} tracks object instances interested in a video sequence, but these methods require the target information determined in the first frame. Different from them, each RGB-D pair is segmented in geometric and semantic manners to obtain correct boundaries in this paper. Moreover, we build a global sparse semantic map in real-time to maintain 2D consistent semantic segments.    

\subsection{2D-3D segmentation}
3D scenes are usually represented by point clouds since this unstructured type of data is efficient and contains rich geometric information compared with 2D images.
3D ShapeNet~\cite{Wu_2015_CVPR} is one of the first works in this area. Via training a 3D convolutional deep belief network from a ground truth shape database. Inspired by ShapeNet, PointNet~\cite{qi2017pointnet} and PointNet++~\cite{qi2017pointnet++} exploit a more efficient representation of 3D surfaces. 
To solve issues in spatio-temporal perception tasks, MinkowskiNet~\cite{choy20194d} making use of 4D dimensional convolutional neural layers outperforms efficient and accurate performance on semantic segmentation. In addition to extracting deep features from point clouds, the researchers also proved that the joint 2D and 3D features complement each other in local areas and obtain better performance.
3DMV~\cite{dai20183dmv} is presented as a joint 3D-multi-view approach built on the core idea of combining spatial and RGB features in an end-to-end architecture. 
Based on 3DMV, 3D-SIS~\cite{hou20193d} fuses 2D color images with 3D geometry features by projecting deep features from 2D RGB view into the voxel grid for instance segmentation.
BPNet~\cite{hu2021bidirectional} enables the bidirectional feature interacting between 2D and 3D CNNs in multiple pyramid levels via the proposed bidirectional projection module. In this paper, after obtaining 2D consistent semantic maps from the system, the SP-Block is proposed to capture features from those 2D semantic segments and transform generated feature volumes to another feature domain encoded from MinkowshiNet~\cite{choy20194d}.

\subsection{Scene understanding from RGB-D sequences}
Based on RGB-D images, tracking and mapping methods~\cite{mur2017orb} are used to build global 3D maps that are bridges for complete scene understanding systems. Multi-feature-based trackers~\cite{Li2021PlanarSLAM,yunus2021manhattanslam} achieve robust estimations in indoor scenes, but those mapping parts aim to maintain sparse features for removing camera pose drift rather than reconstructing dense maps. Different from those methods, KinectFision~\cite{2012KinectFusion} and BundleFusion~\cite{dai2017bundlefusion} focus their mind on dense reconstruction by using GPUs to obtain on-the-fly 3D scene reconstruction. In our system, both tracking and dense mapping are important to be implemented via multi-threads on a CPU, which does not have high requirements for hardware.

Texturing 2D semantic maps to dense maps is a traditional way to build online semantic reconstruction systems. SemanticFusion~\cite{2016SemanticFusion} is a pioneer 3D semantic segmentation system, which fuses semantic surfels labeled by convolutional neural networks incrementally. PanopticFusion~\cite{2020PanopticFusion} extends the 2D-to-3D mapping framework to TSDF voxels where semantic labels come from pixel-wise panoptic prediction networks. Those 2D-to-3D architectures are direct to follow, but they rely on the performances of 2D semantic segmentation methods, which limits further improvement. 
Instead of using 2D semantic results, SceneGraphFusion~\cite{2021SceneGraphFusion} estimates a semantic scene graph based on a 3D instance map while the performance also relies on the instance fusing method.   
Instead of sticking those 2D labels in the dense model or capturing deep features from point clouds only, we try to make the 2D objects' labels consistent and let our 3D segmentation network segment the mesh model in one shoot, which tends to be more accurate since global structures and local details are considered together.

\section{System Overview}
In this section, we introduce the main modules of the pipeline
, as shown in Figure~\ref{fig:pipeline}, the system is divided into three parts, 1) Camera tracker and dense mapping; 2) Semantic label propagation; 3) Two-branch semantic segmentation. 

\subsection{Tracking and mapping}
To achieve robust pose estimation performances in the general indoor environments, the camera tracking module used in this paper is our Manhattan-SLAM~\cite{yunus2021manhattanslam} that exploits points, lines, plane features, and spatial constraints between them, including parallel and perpendicular planes. Note that this tracker can be replaced by other tracking methods.  

For the mapping part, we have two main changes proposed. Compared with the surfel map generated by Manhattan-SLAM, 3D objects' semantic labels are obtained from geometric and semantic segments of every keyframe, which are fused into a global sparse semantic map, as shown in Figure~\ref{fig:class_img}(e), to support consistent 2D semantic mask generation. Furthermore, a dense mapping module is created to propose a smooth dense mesh map. Since sparse maps cannot provide enough information for robots, our system generates a dense mesh map incrementally based on CPU. When a new keyframe is generated from the tracking thread, we make use of the estimated camera pose and the RGB-D pair to build a dense TSDF~\cite{zhou2013dense,niessner2013real} model. After that, the marching cubes method~\cite{lorensen1987marching} is exploited to extract the smooth surface from voxels.

\subsection{2D Consistent semantic segmentation}
The consistent semantic segmentation strategy in this architecture is also an important module that is responsible to provides stable 2D semantic instance predictions of different views. In this module, two segmentation branches, learned~\cite{yolact-iccv2019} (Figure~\ref{fig:class_img}(b)) and geometric~\cite{tateno2015real} (Figure~\ref{fig:class_img}(c\&d)) methods, are used to deal with RGB and depth maps, respectively. The geometric part segments objects' areas based on the 3D shape by computing normals from depth maps, while the learned one predicts the semantic instance masks directly after training convolutional neural networks on large datasets. To remove the boundary noise of semantic labels, only union regions of those two maps are accepted by the system. 

As we all know, images that only capture part information of objects are useful for 3D semantic segmentation since more details and boundaries information can be obtained from there. Those partial scans, however, bring huge challenges to 2D semantic segmentation networks. To keep the consistency of segments, we take advantage of camera poses and the global semantic sparse map to correct those ill-posed results.   
 
\subsection{2D-3D semantic segmentation network}
In this module, an encoder-decoder network is implemented for the final 3D dense semantic segmentation task. In the encoder module, the proposed SP-Block is connected with the original encoder of MinkowskiNet to build deep embedding features that are decoded in semantic predictions.

Benefiting from the SP-Block, deep features from 2D and 3D domains can be fused to predict 3D semantic segments from the 3D dense reconstruction. 
\begin{figure}
    \centering
    \subfigure[\textit{Couch} and \textit{Chair} predicted from Yolact~\cite{yolact-iccv2019}]{
    \includegraphics[width=0.22\linewidth]{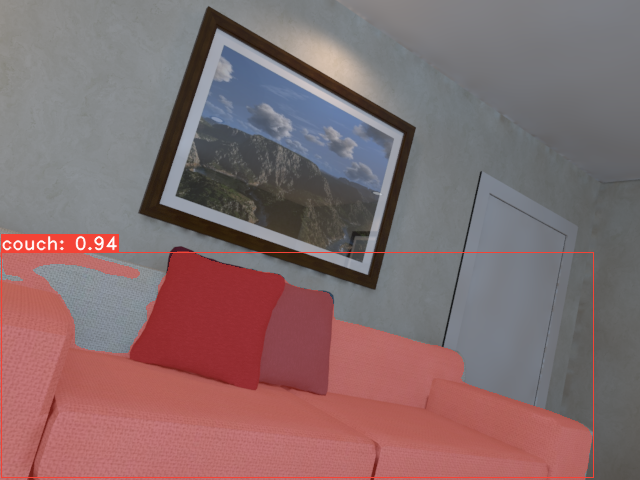}
    \includegraphics[width=0.22\linewidth]{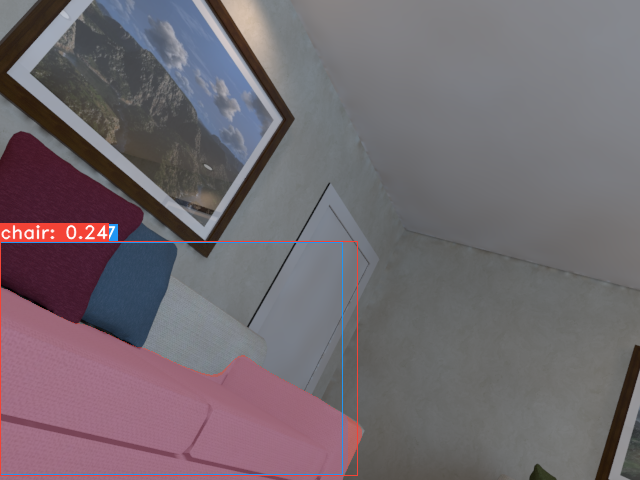}
    }
    \subfigure[\textit{Couch} and \textit{Couch} generated from our method]{
     \includegraphics[width=0.22\linewidth]{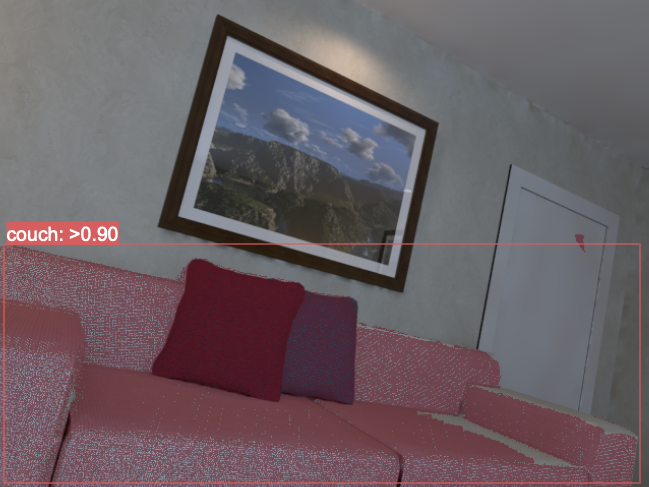}
    \includegraphics[width=0.22\linewidth]{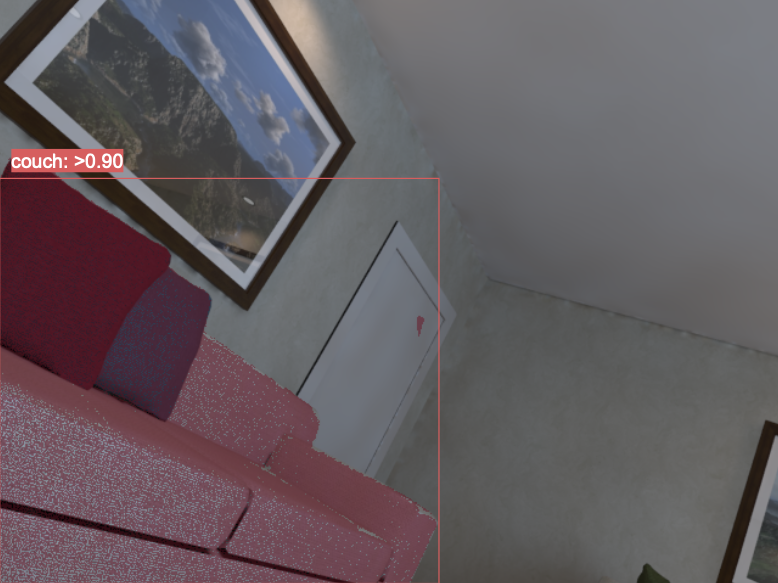}}
    \subfigure[\textit{TV} and \textit{Oven} predicted from Yolact~\cite{yolact-iccv2019}]{
    \includegraphics[width=0.22\linewidth]{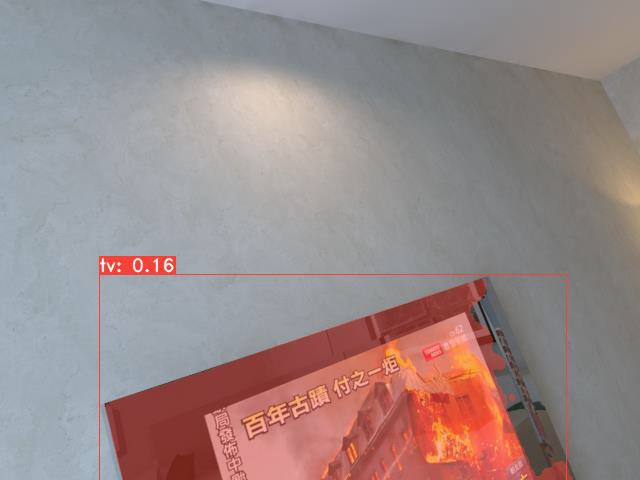}
    \includegraphics[width=0.22\linewidth]{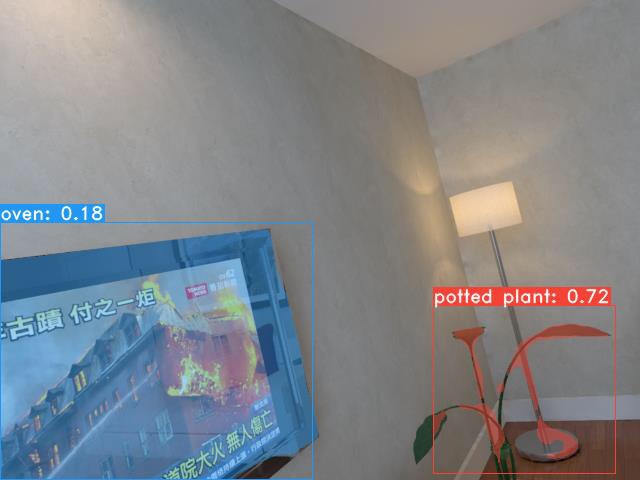}
    }
    \subfigure[\textit{TV} and \textit{TV} generated from our method]{
     \includegraphics[width=0.22\linewidth]{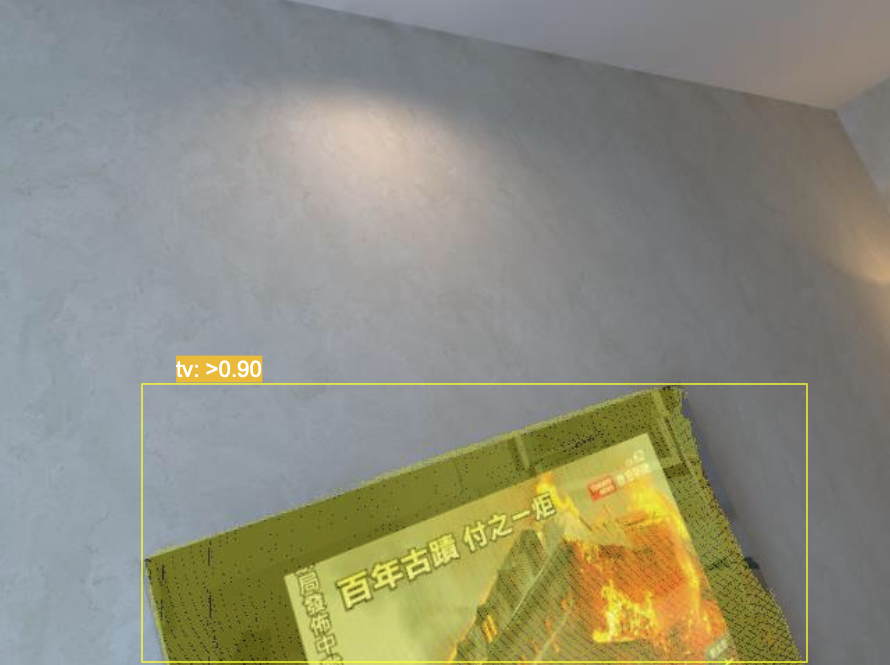}
    \includegraphics[width=0.22\linewidth]{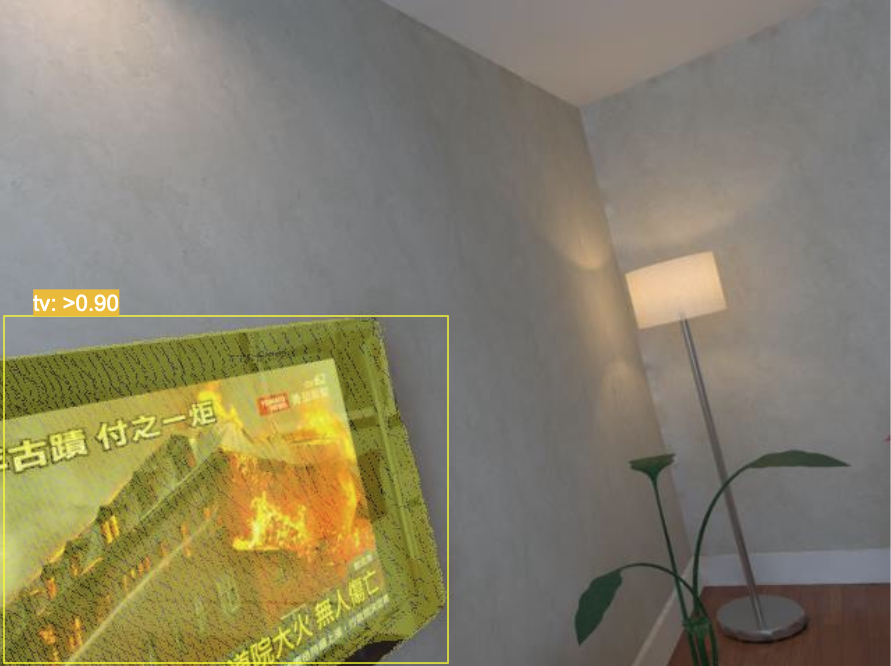}}
    \caption{Consistent semantic segmentation performances in the ICL sequence .}
    \label{fig:my_label}
\end{figure}

\section{Odometry based consistent 2D semantic segmentation} \label{2d semantic}
Inconsistent semantic segmentation prediction between different RGB images of the same scene is a common issue in semantic segmentation methods~\cite{yolact-iccv2019,yolact-plus-tpami2020}. To solve this issue, an incremental joint 2D segmentation strategy is proposed to achieve sharp and consistent segments from each keyframe.

\subsection{ Segments from a single RGB-D pair }
\begin{figure}
    \centering
    \subfigure[Scenes ]{\includegraphics[width=0.23\linewidth]{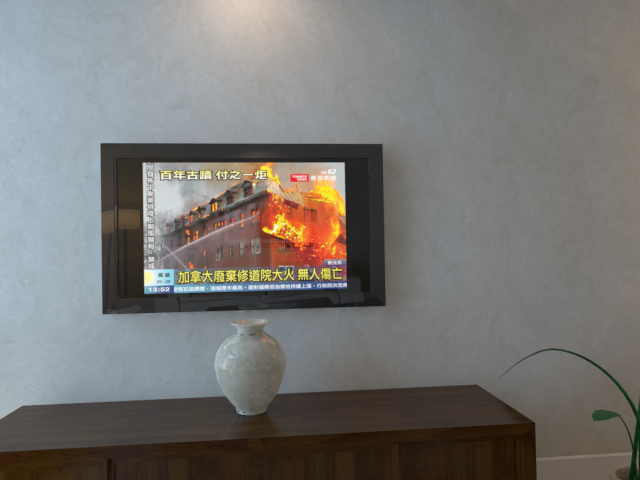}}
    \subfigure[Geometric seg. map]{\includegraphics[width=0.23\linewidth]{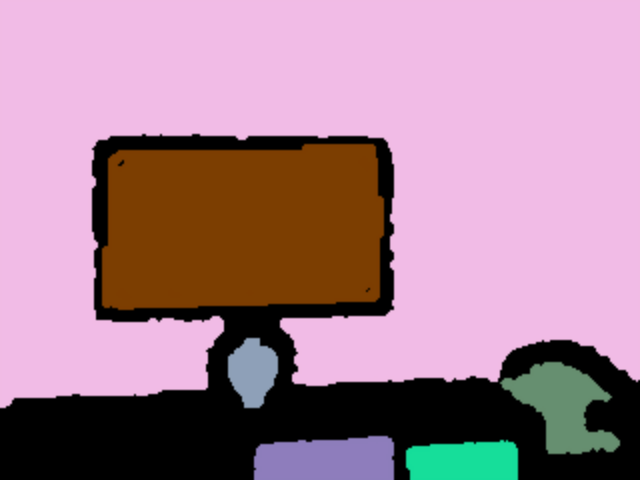}}
    \subfigure[Semantic seg. map map]{\includegraphics[width=0.23\linewidth]{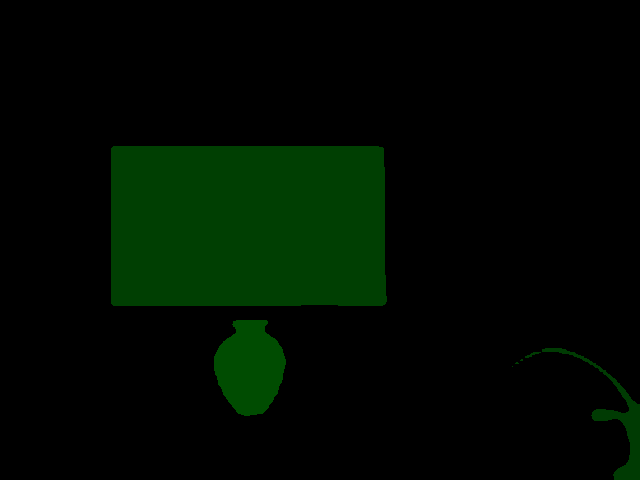}}
    \subfigure[Probability map]{\includegraphics[width=0.23\linewidth]{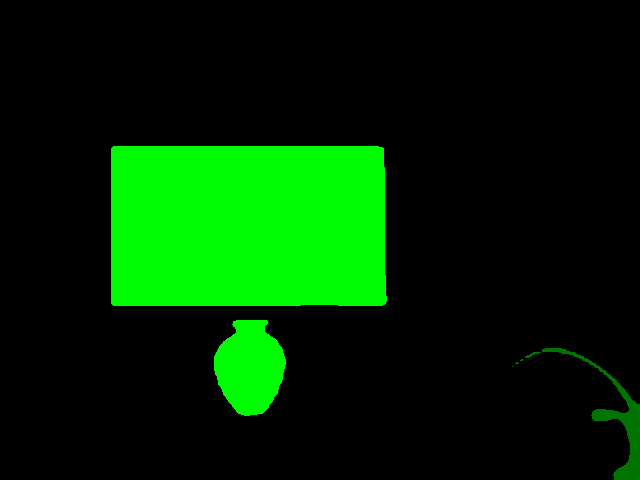}}
    \subfigure[Sparse semantic instance map]{
    \includegraphics[width=0.92\linewidth]{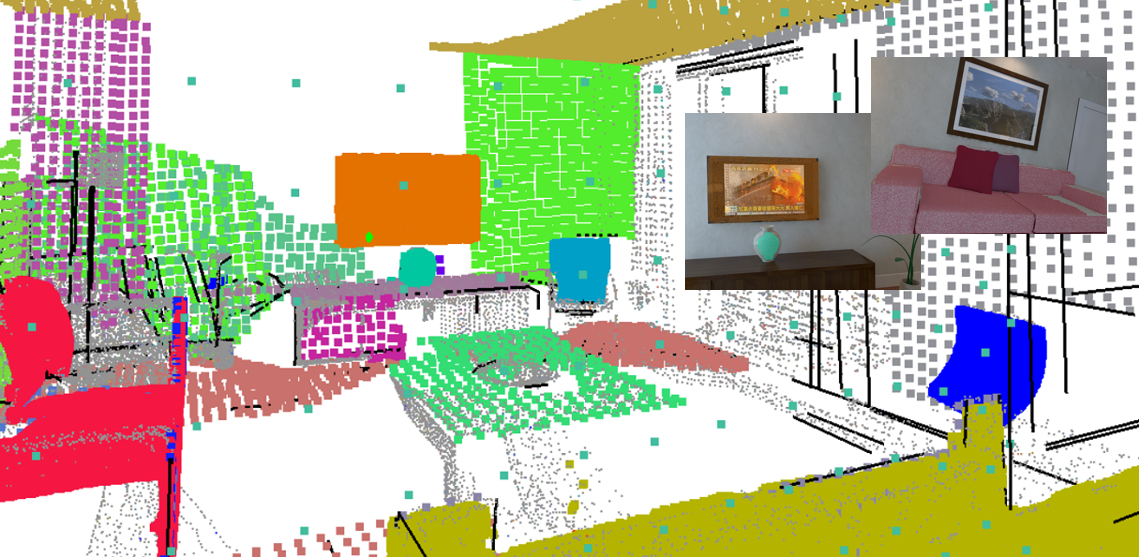}}
    \caption{Sparse semantic instance map building and consistent 2D semantic image generation.}
    \label{fig:class_img}
\end{figure}

In this paper, each RGB image is fed to YOLACT~\cite{yolact-iccv2019} to segment instances and predict objects' labels, there are two types of outputs, label $R_{rgb}$ and probability maps $R_{p}$, from the network, where the first one codes each index of detected objects as shown in Figure~\ref{fig:class_img}(a) while the three channels of the second map (see Figure~\ref{fig:class_img}(b)) is used to save the corresponding probabilities. 


%
Since boundaries of semantic masks generated from an RGB image are commonly noisy, we extract areas with discontinuous depth information from the corresponding depth map. 
Given depth maps, geometric-based shape segmentation methods~\cite{8594391,tateno2015real} are used to segment the scene into different instances according to the normal edge analysis. As shown in Figure~\ref{fig:class_img}, the TV is segmented from a wall since the normal map detects disconnection regions between them. Therefore, a filtered segmentation map $R^*$ is obtained by 
\begin{equation}
    R^* = R_{rgb}\cdot R_{d}
\end{equation}
here $R_{d}$ is a binary map where instance-covered pixels are denoted as 1. We have to note that those segments extracted from the RGB image will be removed if they do not exist in the geometric map, which will be completed when the information appears in both semantic and geometric images. 

\subsection{Semantic propagation}
To keep the consistency of objects captured in different views, we maintain a semantic sparse point clouds map containing different geometric landmarks~\cite{yunus2021manhattanslam,Li2021PlanarSLAM} and semantic objects. 

Each 3D object of the map needs to be initialized and updated in the whole tracking process.  
First, 3D objects $O_i, i\in (1,n)$ where $n$ is the number of 3D objects saved in the map are re-projected to a new keyframe to obtain 2D re-projections regions $o_i^{rp}$. We compute the IoU (Intersection over Union) between $o_i^{rp}$ and semantic labels $o_j, j\in (1,m)$, here $m$ is the number of detected objects of $R^*$. If the IoU is more than a threshold $t_{iou}=0.4$, we continue to check the probability of $o_j$. $o_i^{rp}$ and $o_j$ are matched when index and probability of $o_j$ are matched with $o_i^{rp}$ at the same time. Otherwise, we continually check the probabilities of semantic labels in $R^*$, if they are more than the threshold $t_{p1} = 0.9$, those objects will be noted as new comes and be fused into the map. 

In the objects' fusion process, when the re-projected information is matched with the current keyframe's semantic labels, we will also update the probability of $O_i$. If the probabilities of those 2D semantic segments are more than $t_{p1}$, the probability of $O_i$ will be increased since the object is reconfirmed in different views. If those new segments that satisfy the IoU requirement are in high probabilities but in different semantic labels, the weights of related 3D objects $O_i$ will be decreased, which will be removed from the map when the weights are less than $t_{p2}=0.7$.      

\section{2D-3D joint semantic segmentation}
\begin{figure}
    \centering
    \includegraphics[width=\linewidth]{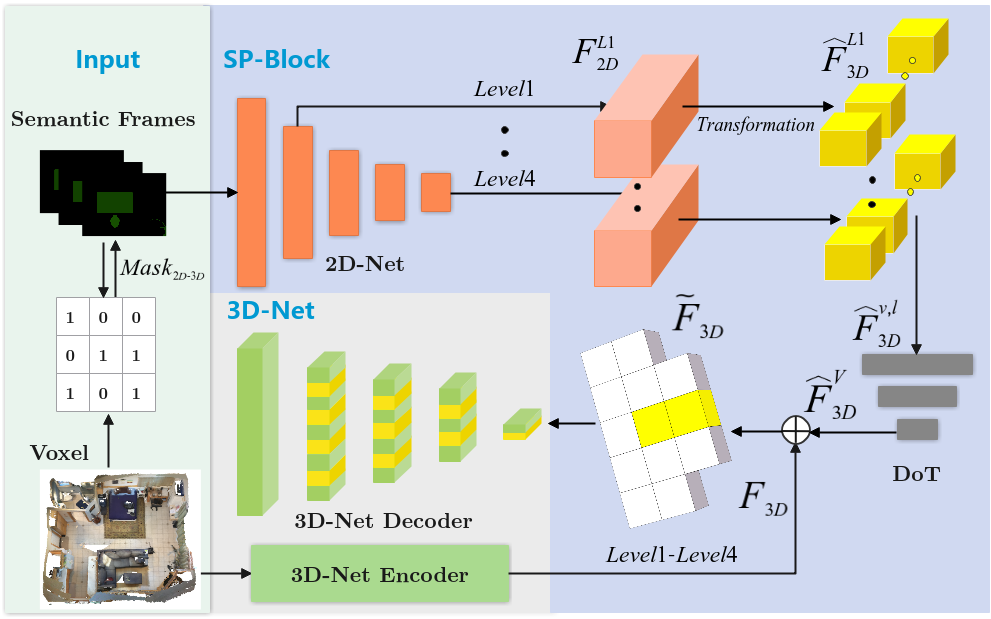}
    \caption{2D-3D semantic segmentation network. The input including 2D semantic segmentation is associated with 3D voxels, and the corresponding relationship mask is established through back projection. 
    }
    \label{fig:FSmodule}
\end{figure}

Given consistent 2D semantic images, camera poses, and a dense 3D model within the same coordinate, an encoder-decoder architecture as shown in Figure~\ref{fig:FSmodule} is introduced in this section to predict semantic segmentation. 

\subsection{Interested regions selection}
3DMV~\cite{dai20183dmv} and 3D-SIS~\cite{hou20193d} propose differentiable projection layers mapping 2D features to corresponding 3D voxels. However, this operation brings noisy to the 3D branch if features from 2D origins have no matches in 3D models. Therefore, we  re-project~\cite{hu2021bidirectional} each 3D point $P=[x, y, z]$ of the model to 2D images to detect intersections that are represented into binary masks $B_k$ by using, 
\begin{equation}
    B_k = KT_{kw}P \ocircle R_k^*
\end{equation}
where $\ocircle$ is the $and$ operation between the $k^{th}$ semantic input image $R_k^*$ and the re-projected image from the 3D model. $T_{kw}$ is the 6 DoFs pose matrix from the world to the $k^{th}$ camera coordinate and $K$ is the intrinsic matrix.

\subsection{Semantic projection block}

As shown in Fgiure~\ref{fig:FSmodule}, the embedding $\widetilde{F}_{3D}$ is constructed by two branches, $F_{3D}$ from the encoder of MinkowskiNet~\cite{choy20194d} and $\hat{F}_{3D}^V$ from our semantic projection block introduced in this section.

First, we extract deep feature pyramids by using ResNet-18~\cite{he2016deep} from multi-view semantic images. There are four levels of feature maps $F_{2D}^{Li}, i\in[1\dots4]$ extracted from each image. To maintain compatibility with deep features from the encoder of 3D-Net~\cite{choy20194d}, we project each feature channel of $F_{2D}^{Li}$ to the shape of $N\times1$ based on the camera pose and intrinsic matrix, where $N$ is the number of voxels in the 3D model. Therefore, each level's feature maps (with $C$ channels) are transferred to a shape as $N\times C$.  
And then at the same level of $V$ views, following~\cite{hu2021bidirectional} we concatenate those transferred shapes along the channel direction to obtain $\hat{F}_{3D}^{v,l}$ with the size of $N\times C\times V$.


Then the DoT operation that is constructed by four 3D sparse convolutional layers and a sparse max-pooling layer is proposed to aggregate feature volumes $\hat{F}_{3D}^{v,l}$ from different views.  
\begin{equation}
    \hat{F}_{3D}^{V} =  \sum_{v=1}^{V}\sum_{l=1}^{L}DoT( \hat{F}_{3D}^{v,l})
\end{equation}
where $L$ is the size of feature levels while $V$ is the number of views. Via the DoT operation, the shape of $\hat{F}_{3D}^{v,l}$ is transferred with the same size of $F_{3D}$. Moreover, the spatial and semantic information from different views is fused.       




Finally, corresponding levels of encodes' features $F_{3D}$ based on MinkowskiNet and $\hat{F}_{3D}^V$ based on our SP-Block are fused by a concatenation operation as 
  \begin{equation}
  \widetilde{F}_{3D} = F_{3D} \oplus \hat{F}_{3D}^V
\end{equation}
where $\widetilde{F}_{3D}$ is the fusion embedding that is fed to the decoder to obtain semantic prediction results.

%


\subsection{Decoder and training}
The decoder used in this paper comes from MinkowskiNet~\cite{choy20194d}, which generates a semantic label for every 3D point. Readers can refer to MinkowskiNet~\cite{choy20194d} for details. Furthermore, we make use of the conventional cross entropy~\cite{2015Deep} to supervise the whole encoder-decoder pipeline. During the raining process, we exploit the ground truth 2D semantic labels and 3D reconstruction models of each scenario as the input of our network.

\subsection{Implementation details}

The network is implemented on the PyTorch platform, which exploits the SGD optimizer with a learning rate of $0.01$ and a momentum of $0.9$. Furthermore, the network is trained on the machine with 4 NVIDIA GeForce 2080TI GPUs and 64GB RAM, where the batch size is set to 12 in 100 training epochs.

The size of RGB-D images fed to our tracking system is $480\times 640$, while the size of semantic images for SP-Block is downsampled to $240\times 320$. The channels' size of four layers in the feature pyramid are $512$, $256$, $128$ and $96$, respectively. After the DoT operation, the channels' size of four layers in $\hat{F}_{3D}^V$ are $256$, $128$, $128$ and $96$, respectively.

\begin{figure*}[]
\centering
\subfigure[Dense Map]{
    \begin{minipage}[t]{0.22\linewidth}
        \centering
        \includegraphics[width=\linewidth]{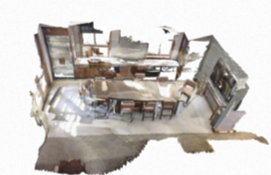}\\
        \vspace{0.02cm}
        \includegraphics[width=\linewidth]{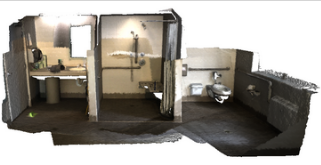}\\
        \vspace{0.02cm}
        \includegraphics[width=\linewidth]{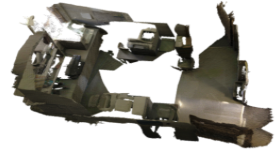}\\
        \vspace{0.02cm}
    \end{minipage}%
}%
\subfigure[MinkowskiNet~\cite{choy20194d}]{
    \begin{minipage}[t]{0.23\linewidth}
        \centering
        \includegraphics[width=\linewidth]{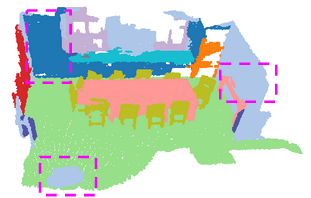}\\
        \vspace{0.02cm}
        \includegraphics[width=\linewidth]{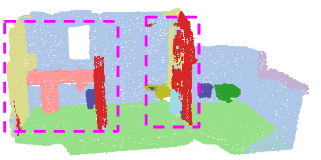}\\
        \vspace{0.02cm}
        \includegraphics[width=\linewidth]{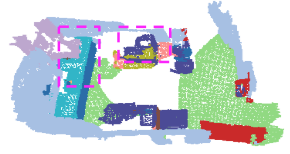}\\
        \vspace{0.02cm}
    \end{minipage}%
}%
\subfigure[Ours]{
    \begin{minipage}[t]{0.235\linewidth}
        \centering
        \includegraphics[width=\linewidth]{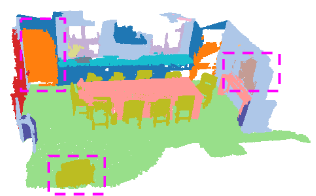}\\
        \vspace{0.02cm}
        \includegraphics[width=\linewidth]{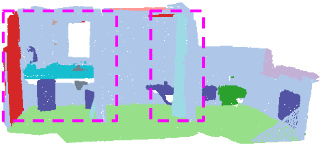}\\
        \vspace{0.02cm}
        \includegraphics[width=\linewidth]{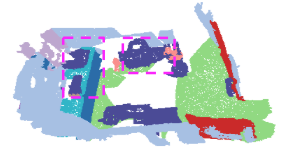}\\
        \vspace{0.02cm}
    \end{minipage}%
}%
\subfigure[Ground Truth]{
    \begin{minipage}[t]{0.23\linewidth}
        \centering
        \includegraphics[width=\linewidth]{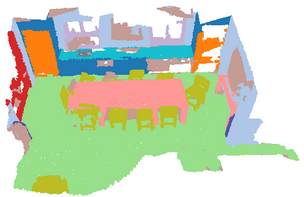}\\
        \vspace{0.02cm}
        \includegraphics[width=\linewidth]{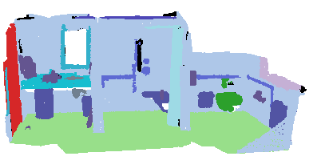}\\
        \vspace{0.02cm}
        \includegraphics[width=\linewidth]{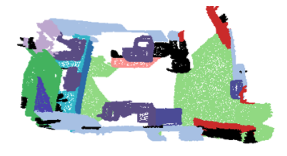}\\
        \vspace{0.02cm}
    \end{minipage}%
}%

\includegraphics[width=\linewidth]{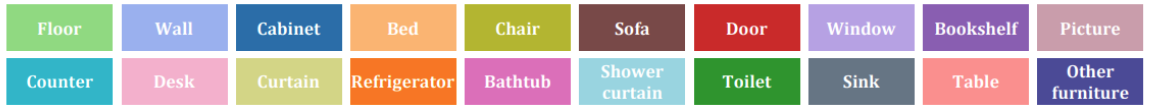}
\centering
\caption{A qualitative comparison of 3D semantic segmentation between MinkowskiNet and our method. We use pink boxes to highlight the difference between them. Different classification information is represented by different colors, and the classification information corresponds to the bottom color palette. }
\vspace{-0.2cm}
\label{fig:compare_fig}
\end{figure*}

\section{Experiments}
In this section, the performances, including dense mapping and 3D segmentation, of the system are evaluated on public datasets and compared with state-of-the-art methods.  
\subsection{Dataset}
\subsubsection{ScanNetV2}

The ScanNetV2~\cite{2017ScanNet} dataset includes 1513 sequences (around 2.5 million RGB-D frames) from 70 unique indoor scenes, which provides ground truth annotations for training, validation, and testing directly on 3D reconstructions. Those sequences are split into training, validation, and testing datasets where the semantic labels are defined according to the rule of NYU40~\cite{Silberman:ECCV12}. 
\subsubsection{3RScan}
The 3RScan dataset~\cite{wald2019rio} is a large indoor RGB-D dataset scanned multiple times of changing environments. It contains 1482 RGB-D scans of 478 environments and ground truth annotations of instance-level semantic segmentation, dense mapping, scene semantic segment.
\subsubsection{ICL-NUIM}
The ICL-NUIM synthetic dataset aims at benchmarking tracking and mapping methods, which provides the living room and the office room scene with ground-truth poses. And the ground-truth maps are only provided for 4 sequences of the living room. 


\begin{table}[]
\centering
\scalebox{0.87}{
\begin{tabular}{l|cccc}
\toprule
     & ElaticFu\cite{whelan2016elasticfusion}  &BundleFu\cite{dai2017bundlefusion} & InfiniTAM\cite{prisacariu2017infinitam}  & Ours \\ \hline
lr2 & 0.8 &0.7 & 0.1   &0.7      \\ 
lr3 & 2.8  & 0.8   & 2.8 &0.7      \\ 
\bottomrule
\end{tabular}}
\caption{Reconstruction error (cm) for the Living Room sequences.}
\end{table}

\subsection{Accuracy of dense reconstruction}
First, the reconstruction accuracy is evaluated between different approaches. ElasticFusion~\cite{whelan2016elasticfusion}, BundleFusion~\cite{dai2017bundlefusion} and InfiniTAM~\cite{prisacariu2017infinitam} are dense reconstruction methods, where the first one reconstructs surfel-based models and the others obtain dense mesh model as well as ours. In the lr2 and lr3 sequences, a complete room can be found. Therefore, we compare the reconstruction errors among those methods. Compared with those methods, our method is more robust in different sequences and we only need a CPU for the on-the-fly dense reconstruction.

\begin{figure}
    \centering
    \subfigure[Dense reconstruction]{
    \includegraphics[width=0.48\linewidth]{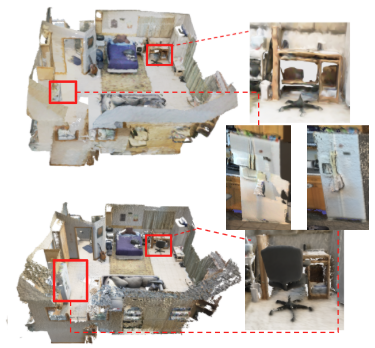}}
    \subfigure[Semantic segmentation]{
    \includegraphics[width=0.45\linewidth]{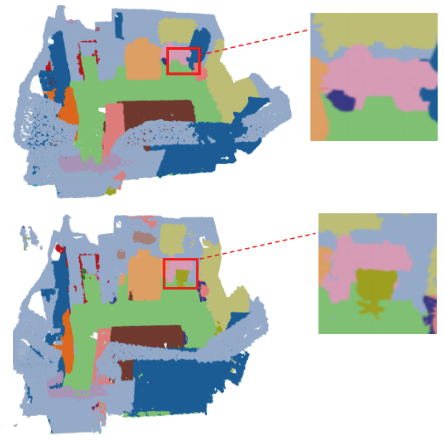}}
    \caption{Dense reconstruction results. Up: ScanNet ground truth. Down: ours.}
    \label{fig:bundle_ours}
\end{figure}

Moreover, we compare the qualitative reconstruction results between ours and the ground truth of ScanNetv2 that is built from BundleFusion. As shown in Figure~\ref{fig:bundle_ours}(a), our method can reconstruct the chair completely. Benefiting from our accurate pose estimation module and the smooth dense reconstruction strategy, the refrigerator is reconstructed more accurately than the ground truth. Related semantic segmentation results are shown in Figure~\ref{fig:bundle_ours}(b).

\begin{table*}[]
\setlength{\arraycolsep}{0.5pt}
\scalebox{0.71}{
\begin{tabular}{l|lccccccccccccccccccc}
Method      & bath & bed & bkshf & cab & chair & cntr & curt & desk & door & floor & other & pic & fridge & shower & sink & sofa & table & toilet & wall & window \\ \hline
SF~\cite{2016SemanticFusion}
&59.8&	46.8&	32.0&	35.7&	46.9&	33.2&	46.9&	34.7&	35.7&	72.2&	34.7&	21.7&	34.3&	28.8&	47.2&	43.7&	37.8&	65.5&	58.3&	29.5     \\

SR~\cite{2018Semantic}
&69.7&	52.6&	31.2&	31.7&	64.0&	24.0&	30.3&	26.1&	30.9&	80.6&	33.3&	7.3&	56.3&	23.6&	46.2&	58.3&	51.6&	73.3&	66.9&	21.1      \\
PF~\cite{2020PanopticFusion}
&57.5&	67.0&	48.4&	44.8&	66.7&	35.8&	53.3&	42.0&	35.6&	81.0&	40.3&	30.2&	47.0&	50.8&	52.6&	61.3&	54.8&	82.1&	65.7&	45.9
   \\
PsF~\cite{Quang0Real}
&65.6&	61.2&	65.7&	48.6&	68.4&	41.7&	54.9&	48.9&	47.5&	87.1&	43.7&	25.7&	41.8&	34.5&	53.4&	59.8&	54.0&	78.9&	70.6&	47.0   \\
FPC~\cite{2020Fusion}
&\textbf{85.4}&\textbf{82.3}&	6.4&	60.9&	75.1&	56.0&	\textcolor{blue}{64.8}&	58.2&	\textbf{64.8}&	91.9&	46.4&	\textbf{40.6}&	\textbf{64.2}&	51.7&	\textcolor{blue}{63.5}&	77.9&	68.9&	87.0&	\textbf{83.8}&	56.3     \\

SPV~\cite{huang2021supervoxel}  &73.4   &   78.5     & \textbf{79.1}    & 60.5    & 80.6        & \textcolor{blue}{59.3}      & \textbf{70.4}       & 59.9        & \textcolor{blue}{60.5}       & 91.1       & \textbf{57.8}     &  \textcolor{blue}{35.0}        &\textcolor{blue}{57.5}       &\textbf{75.2}  &  61.3      & 72.6        &  64.4        &  86.4      &80.5       & \textbf{61.7}  \\ 
BPNet~\cite{hu2021bidirectional} &83.0&	\textcolor{blue}{80.1}&	\textcolor{blue}{78.2}&	\textbf{61.8}&	\textbf{89.0}&	\textbf{61.9}&	58.5&	\textbf{65.7}&	57.1&	\textcolor{blue}{93.8}&	\textcolor{blue}{53.3}&	23.7&	44.2&	61.7&	\textbf{65.2}&	\textcolor{blue}{79.4}&	\textbf{72.7}&	\textbf{89.5}&	\textcolor{blue}{81.7}&	\textcolor{blue}{59.3} \\ \hline
Ours	& \textcolor{blue}{83.4}&	79.2&	76.9&	\textcolor{blue}{61.0}& \textcolor{blue}{88.9}&	57.9&	58.3&	\textcolor{blue}{64.2}&	58.3&	\textbf{93.9}&	53.2&	24.3&	44.4&	\textcolor{blue}{65.1}&	63.4&	\textbf{80.2}&	\textcolor{blue}{71.4}&	\textcolor{blue}{89.0}&	81.4&	58.5    \\ 
Ours+& 84.9&	81.2&	80.2&	65.5&	89.9&	61.4&	59.1&	70.5&	58.7&	93.9&	58.0&	25.6&	48.3&	62.3&	65.9&	82.8&	75.5&	88.5&	82.6&	61.0  \\ \hline
\end{tabular}}
\caption{The Quantitative Accuracy Comparison of the Final Semantic Segmentation Results on the ScanNetV2 Validation dataset. We use bold and blue numbers to mark the best and second results per instance, respectively.}
\label{tab:camper}
\end{table*}

\begin{table}[]
\scalebox{1}{
\begin{tabular}{l|c|ccc}
Method   & recon. &  voxel    & mIoU & mAcc  \\ \hline
SF~\cite{2016SemanticFusion}$^*$  & Y&- &42.2&47.4     \\
SR~\cite{2018Semantic}$^*$ & Y&- & 44.0 & 65.6\\
PF~\cite{2020PanopticFusion}$^*$ & Y& - & 53.1 & 68.7 \\
PsF~\cite{Quang0Real}$^*$ & Y&-&55.0 &70.3  \\
FPC~\cite{2020Fusion}$^*$ &Y &- &67.2 &77.0 \\
SPV~\cite{huang2021supervoxel}$^*$  &Y & 1cm &68.3 &	79.6 \\
BPNet~\cite{hu2021bidirectional} & N    &5cm  &67.1&88.3  \\  \hline
Ours      & Y & 5cm &   67.4     & 84.0    \\
Ours+    & Y   & 5cm&   69.8    & 88.8    \\ \hline
\end{tabular}}
\caption{3D semantic segmentation mIoU and mAccu results on the ScanNetV2 Validation Set. 'Y' meams that a method has a dense reconstruction function. We use '-' to mark unsure situations and '*' means the result coming from~\cite{huang2021supervoxel}.  }
\label{tab:campare2}
\end{table}

\subsection{3D semantic segmentation results}
Following the common evaluation metrics in previous works, the standard mean Intersection over Union (mIoU) and mean Accuracy (mAcc) as used to evaluate the performance of our network.
Our 3D semantic segmentation results are shown in Table~\ref{tab:camper},\ref{tab:campare2}, where we compare our network with state-of-the-art pipelines. Similar to our methods, SF~\cite{2016SemanticFusion}, SR~\cite{2018Semantic}, SPV~\cite{huang2021supervoxel} and PF~\cite{2020PanopticFusion} are semantic reconstruction systems based RGB-D images, while BPNet~\cite{hu2021bidirectional} deals with point clouds.

Semantic segmentation results of 20-class objects/scenarios from different approaches are listed in Table~\ref{tab:camper}, FPC~\cite{2020Fusion} achieves good predictions in $5$ classes, especially in \textit{bath},\textit{bed} and \textit{wall} instances. However, it does not understand bookshelves existed in scenes. Compared with those methods, our network achieves more robust performances. For most object classes, our method obtains better results. To be specific, we obtain the best results at \textit{chair}, \textit{desk}, \textit{sofa} and \textit{toilet} instances since our 2D semantic masks based on Yolact~\cite{yolact-iccv2019} can detect those objects from RGB images, which proves that the proposed SP-Block improves the segmentation results. Similar to our method, BPNet is a 2D-3D joint method. But, it feeds the network with RGB images. According to the results of Ours$+$, which makes use of the 2D semantic masks provided by ScanNetV2 to take place of Yolact's results, the performances of our network are improved shapely and perform better over BP-Net in the majority of categories listed in Table~\ref{tab:camper}. 

Furthermore, we compare the mIoU and mAcc between different methods as shown in Tabel~\ref{tab:campare2}, where BPNet and ours results are based on the voxel size of $5$ cm while SPV makes use of $1$ cm voxels. When the voxel size becomes smaller, more detailed information is saved in models, which tends to produce accurate predictions. However, it brings more intensive computation and higher requirements for hardware.        

\begin{figure}
    \centering
    \subfigure[Scene from RIO]{
    \includegraphics[width=0.61\linewidth]{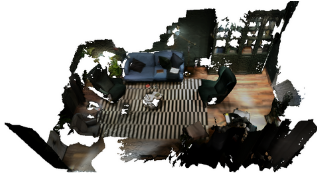}}
    \subfigure[Segmentations]{
    \includegraphics[width=0.33\linewidth]{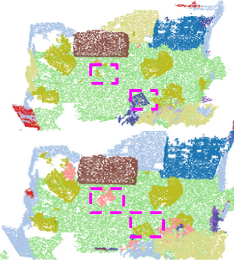}}
    \caption{Results on the RIO dataset.(b)Segmentation results from MinkowskiNet (up) and ours (down)}
    \label{fig:rio}
\end{figure}

To verify the generality of our algorithm, the network is tested on the 3RScan dataset. As shown in Figure~\ref{fig:rio}(b), our approach accurately predicts the table and chair of the scene, but MinkowskiNet fails to detect them.

\section{Conclusion}
In this work, we present a complete scene understanding system starting from RGB-D sequences, which builds a dense mesh map incrementally and segments the map semantically. First, we propose a semantic instance sparse map to support a 2D consistent semantic generation task.  Moreover, the proposed SP-Block is used to extract deep features from those 2D semantic views and project those features to the domain from point clouds. Extensive qualitative and quantitative results show that the proposed method achieves complete and state-of-the-art performances in this area.  

Since the consistent 2D semantic masks can improve the 3D segmentation results, our future research plan is to build a semantic deep visual odometry in an end-to-end architecture, especially focus on a video-based consistent 2D segmentation. 

{\small
\bibliographystyle{IEEEtran}
\bibliography{root}
}

\end{document}